\documentclass[a4paper]{article}

\usepackage{INTERSPEECH2021,amsmath,graphicx,amssymb,bm,multirow,booktabs,hhline,xcolor,subcaption,placeins}

\title{Multitask Training with Text Data for End-to-End Speech Recognition}
\name{Peidong Wang$^{1,2*}$, Tara N. Sainath$^1$, Ron J. Weiss$^1$ \thanks{*This work was performed during an internship at Google.}}
\address{$^1$Google, New York, NY, USA\\
        $^2$The Ohio State University, Columbus, OH, USA}
\email{wang.7642@osu.edu, \{tsainath, ronw\}@google.com}
\begin{document}
%
\maketitle
\begin{abstract}
We propose a multitask training method for attention-based end-to-end speech recognition models. We regularize the decoder in a listen, attend, and spell model by multitask training it on both audio-text and text-only data.
Trained on the 100-hour subset of LibriSpeech, the proposed method, without requiring an additional language model, leads to an 11\% relative performance improvement over the baseline and approaches the performance of language model shallow fusion on the test-clean evaluation set.
We observe a similar trend on the whole 960-hour LibriSpeech training set.
Analyses of different types of errors and sample output sentences demonstrate that the proposed method can incorporate language level information, suggesting its effectiveness in real-world applications.


\end{abstract}
\section{Introduction}
\label{sec:intro}
Attention-based end-to-end (E2E) speech recognition systems map audio features directly to text-level representations \cite{chorowski2015attention,rao2017exploring,chiu2018state,he2019streaming,prabhavalkar2021less}. Various model architectures \cite{chan2016listen,kim2017joint,han2020contextnet,gulati2020conformer,li2021better} and training schemes \cite{cui2018improving,wang2019token,wang2019large} were proposed. The models are typically trained on transcribed speech datasets comprised of parallel audio-text pairs. Such audio-text pairs are more difficult and expensive to obtain compared with audio or text data in isolation. Recently, substantial performance improvements have been made by leveraging audio-only data for speech recognition \cite{baevski2019vq,xu2020iterative,baevski2020wav2vec}.

The most common method for leveraging text-only data is to train a language model (LM) and integrate it into the recognition process using shallow \cite{chorowski2017towards}, cold \cite{sriram2018cold}, or deep fusion \cite{kannan2018analysis}. These methods utilize a second neural network model and thus require additional space and computational resources, making them difficult to deploy in resource-constrained environments such as on-device ASR systems.

Another way to use text-only samples is to convert them to audio-text pairs using text-to-speech synthesis (TTS) techniques. Inspired by the back-translation method in neural machine translation \cite{sennrich2016improving}, Li \emph{et al.} proposed to train the ASR model using audio-text pairs generated from text-only data \cite{li2019semi}. Multiple methods were proposed to jointly train the ASR and TTS models in a cycle-consistent manner \cite{hori2019cycle,baskar2019semi,ren2019almost}. Wang \emph{et al.} used a loss term to encourage the ASR model to generate consistent outputs on real and synthesized presentations of the same utterance \cite{wang2020improving}. These methods face the problem that synthesized audio may bias the ASR model towards unrealistic speech. 

As an alternative to LM fusion and TTS, knowledge distillation methods were proposed to transfer the knowledge in an LM to the ASR model \cite{bai2019learn,futami2020distilling}. An LM is first trained using a large amount of text-only data. To train the ASR model, LM model outputs on the transcripts of the audio-text data are used as soft labels. This approach uses a pre-trained LM during ASR training and does not explicitly incorporate the large amount of text-only data into the ASR model.

The joint acoustic and text decoder (JATD) model \cite{sainath2020attention} incorporates the text transcribed by a conventional ASR model in order to work in two modes, ASR and language modeling.
It is used in a hybrid network designed for two-pass recognition, a transducer for streaming recognition followed by a non-causal rescoring pass using an attention-based decoder \cite{sainath2019two}.
In this study, we propose multitask training with text-only data (MUTE), which differs from JATD in multiple aspects. First, we incorporate reference text directly during training, without using any corresponding audio data and external ASR model. Second, MUTE is designed for one-pass recognition and uses a single attention-based decoder pass during inference.
Compared with JATD, a more closely related area of MUTE is the subtraction of internal LMs for end-to-end ASR models \cite{variani2020hybrid,mcdermott2019density}.
MUTE can be viewed as a method to regularize the internal LM so that it does not overfit smaller training datasets of audio-text pairs. Experimental results on the 100-hour subset of LibriSpeech show that MUTE can effectively incorporate text-only data into E2E models, outperforming the baselines trained using audio-text pairs alone and approaching the performance of LM shallow fusion. We observe a similar trend using the whole 960-hour LibriSpeech training set.

The remainder of this paper is organized as follows. We describe MUTE in Section \ref{sec:sys}. In Section \ref{sec:exp} and \ref{sec:eval}, we present the experimental setup and evaluation results, respectively. Concluding remarks are given in Section \ref{sec:conc}.

\section{System Description}
\label{sec:sys}

\subsection{Attention-Based End-to-End Speech Recognition}
\label{ssec:conventional}
Let us denote the input feature to an attention based end-to-end ASR model as $\boldsymbol{x} \in \mathbb{R}^{T \times F}$ and the corresponding output token sequence as $\boldsymbol{y} \in \mathbb{R}^{U}$, where $F$ is the input feature dimension, and $T$ and $U$ denote the lengths of the input and output, respectively. For audio-text training samples, we denote the output as $\boldsymbol{y}^{a} \in \mathbb{R}^{U^{a}}$, where $a$ refers to audio-text.
A typical E2E system models the following distribution at output token step $u$:
\begin{equation}
    \label{eq:conventional}
    p({y}^{a}_u | \boldsymbol{x}, \boldsymbol{y}^{a}_{1:u-1}; \theta)
\end{equation}
where $\theta$ denotes the model parameters.


\begin{figure*}[thb]
    \begin{subfigure}[b]{.42\textwidth}
        \centering
        \includegraphics[width=.4\linewidth]{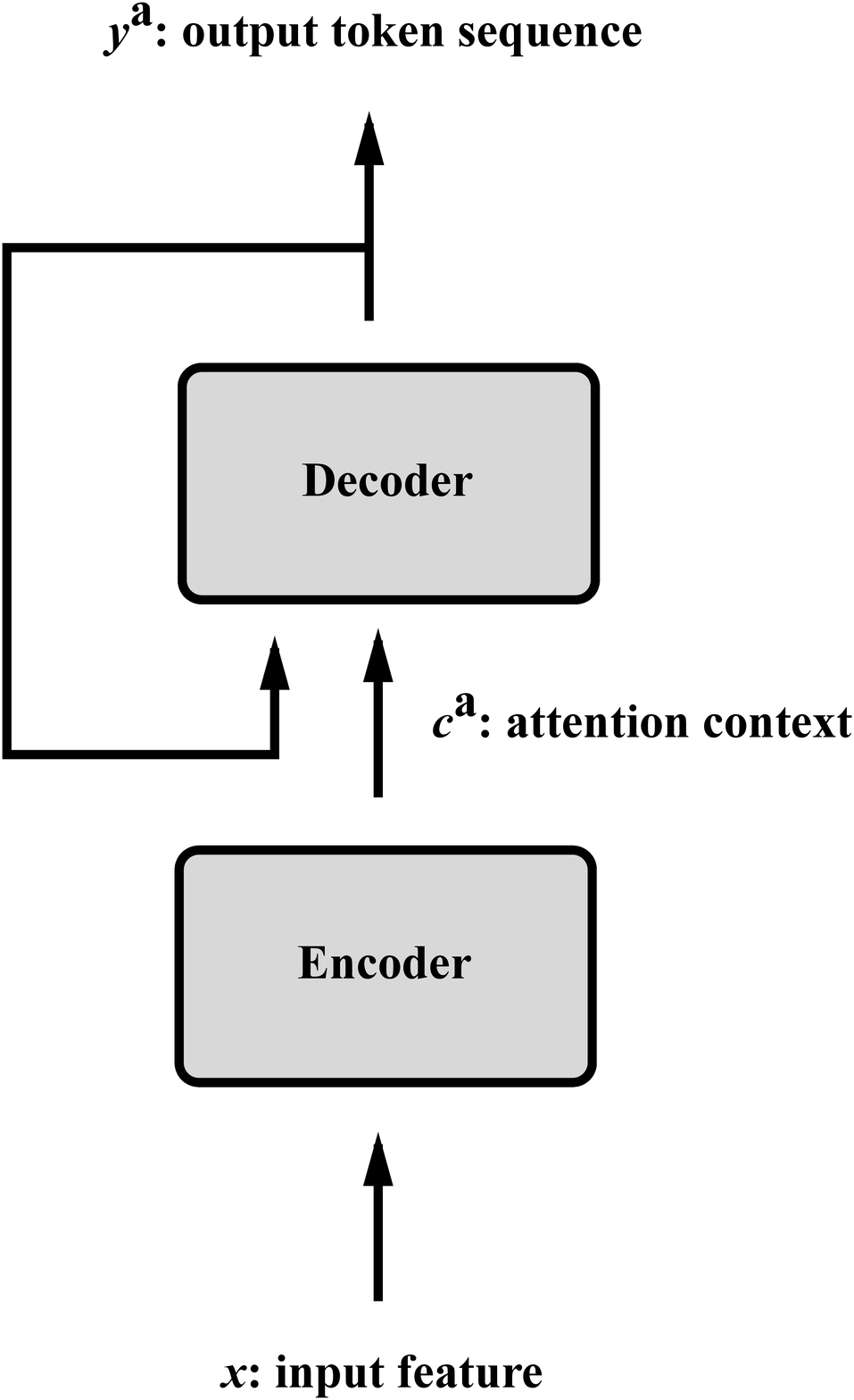}
        \caption{Stage 1}
        \label{fig:mute_stage1}
    \end{subfigure}%
    \begin{subfigure}[b]{.21\textwidth}
      \centering
      \includegraphics[width=.8\linewidth]{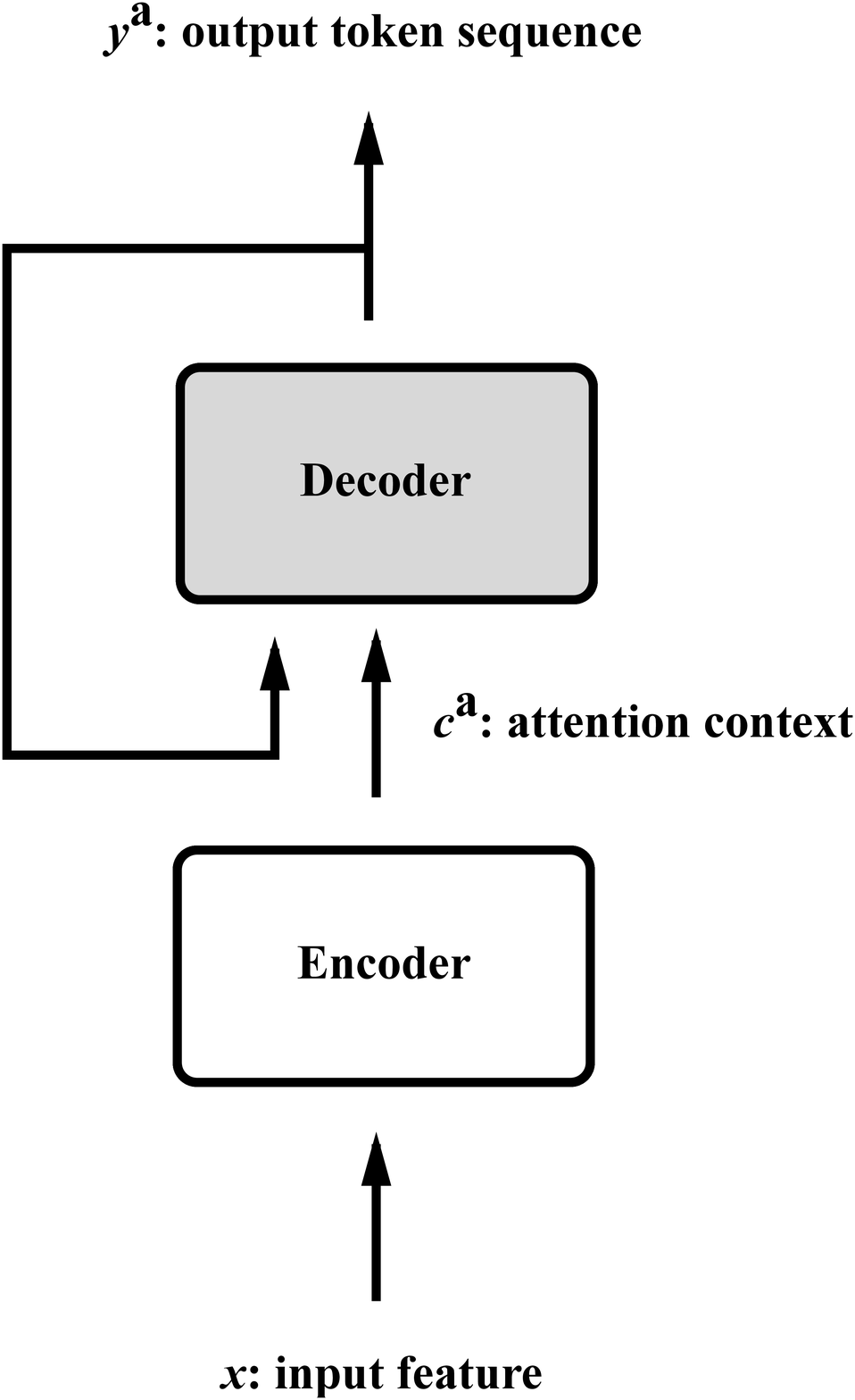}
      \caption{Stage 2: audio-text}
      \label{fig:mute_stage2_audio}
    \end{subfigure}
    \begin{subfigure}[b]{.21\textwidth}
      \centering
      \includegraphics[width=.8\linewidth]{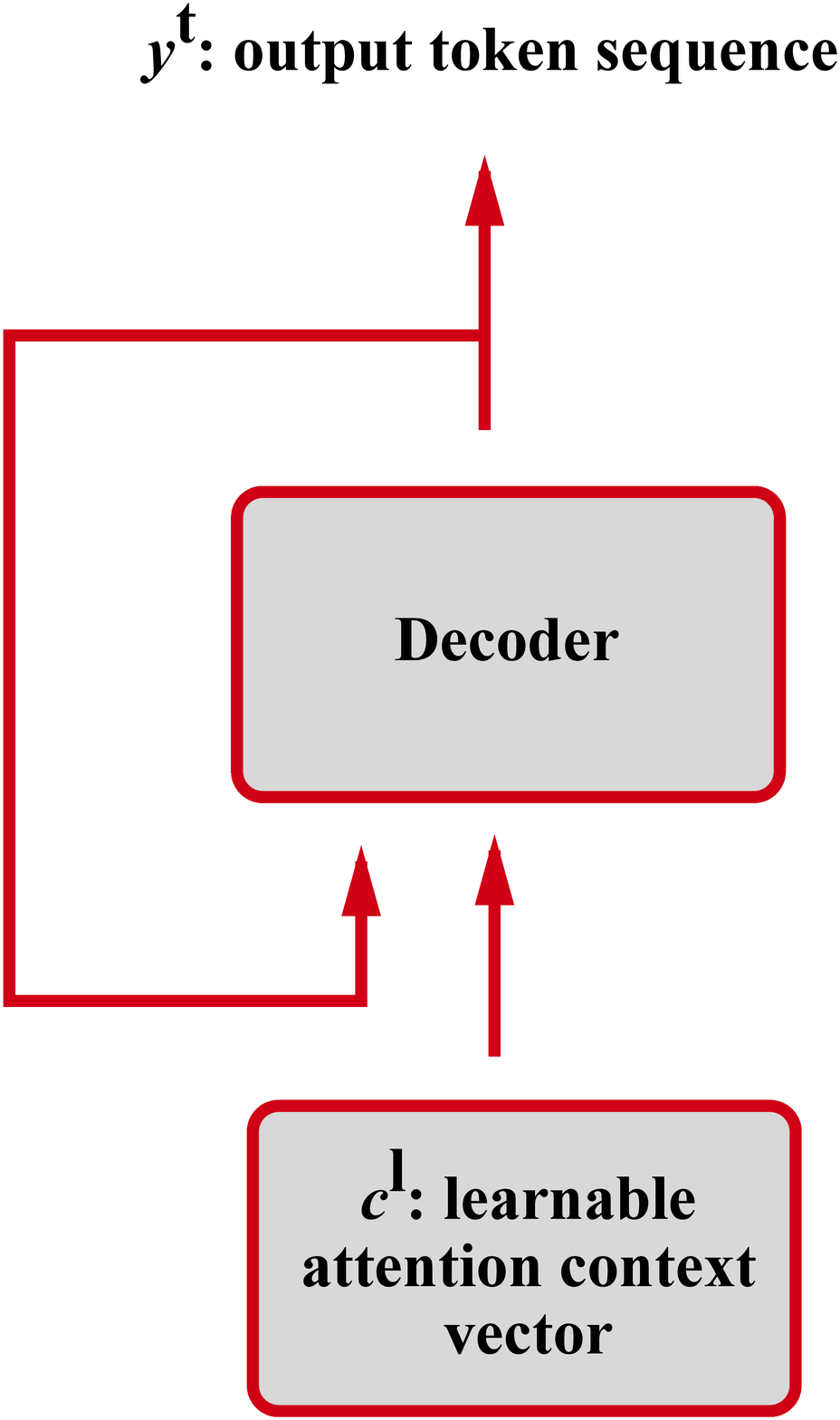}
      \caption{Stage 2: text-only}
      \label{fig:mute_stage2_text}
    \end{subfigure}
    
\centering
 \caption{Illustration of the two stages of MUTE training. Components that are updated during training are in gray, and those which are fixed are white. Black arrows and boxes are used to denote audio-text pairs, while red denotes text-only data.}
 \label{fig:mute_overview}
\end{figure*}

\begin{figure*}[thb]
    \begin{subfigure}[b]{.22\textwidth}
        \includegraphics[width=.6\linewidth]{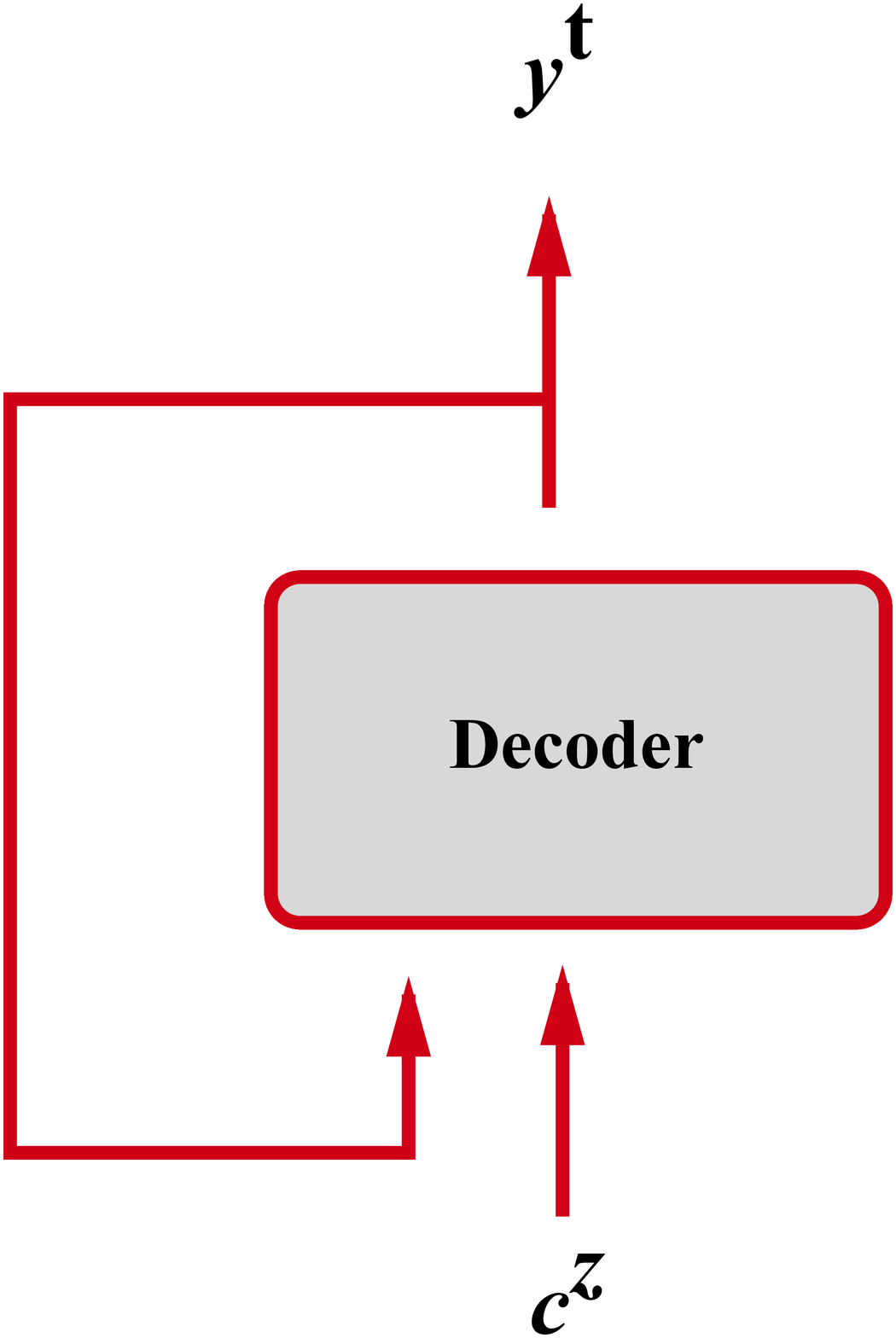}
        \caption{MUTE-Z}
        \label{fig:mutes-a}
    \end{subfigure}%
    \begin{subfigure}[b]{.22\textwidth}
        \includegraphics[width=.6\linewidth]{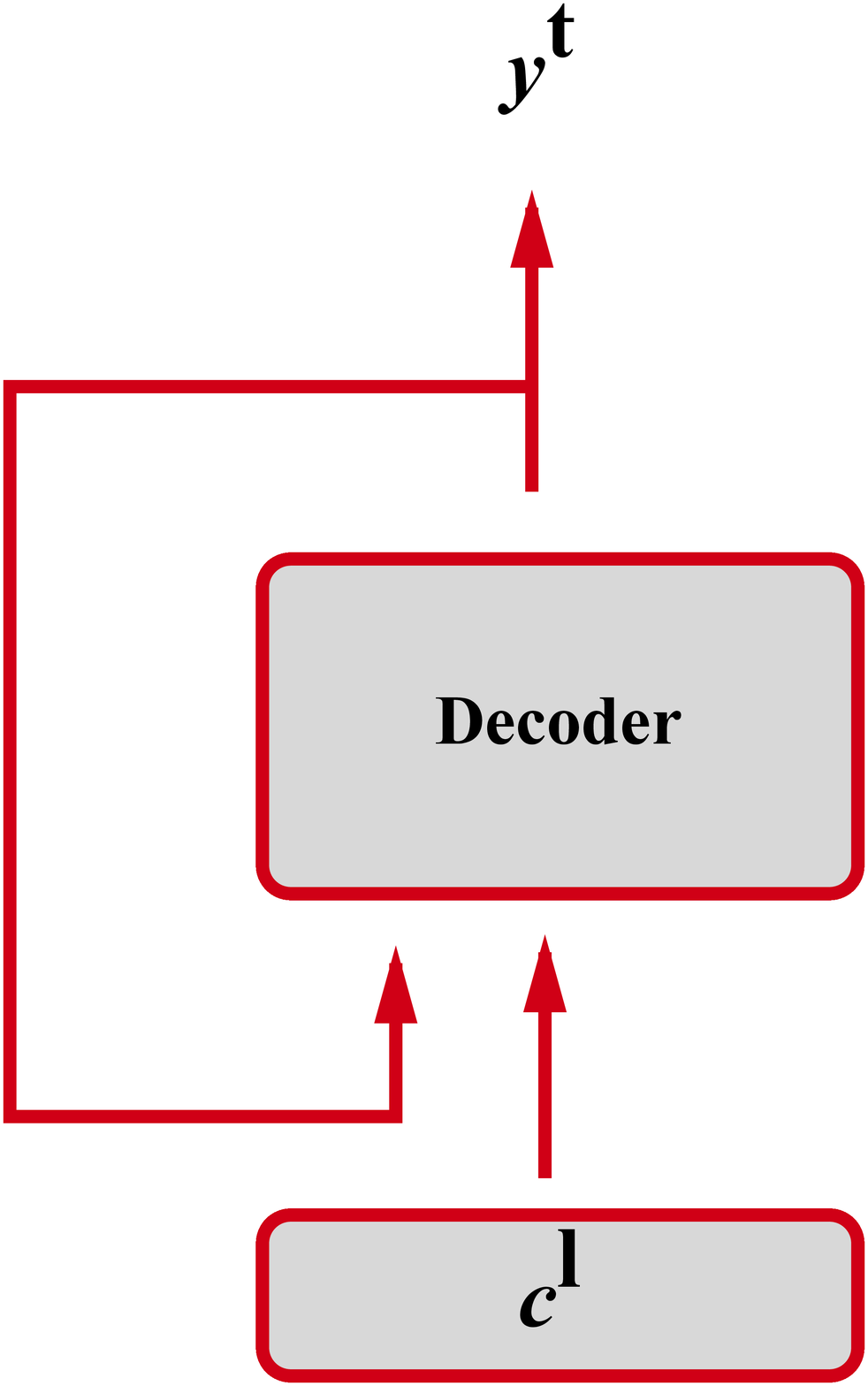}
        \caption{MUTE-L}
        \label{fig:mutes-b}
    \end{subfigure}%
    \begin{subfigure}[b]{.22\textwidth}
        \includegraphics[width=.6\linewidth]{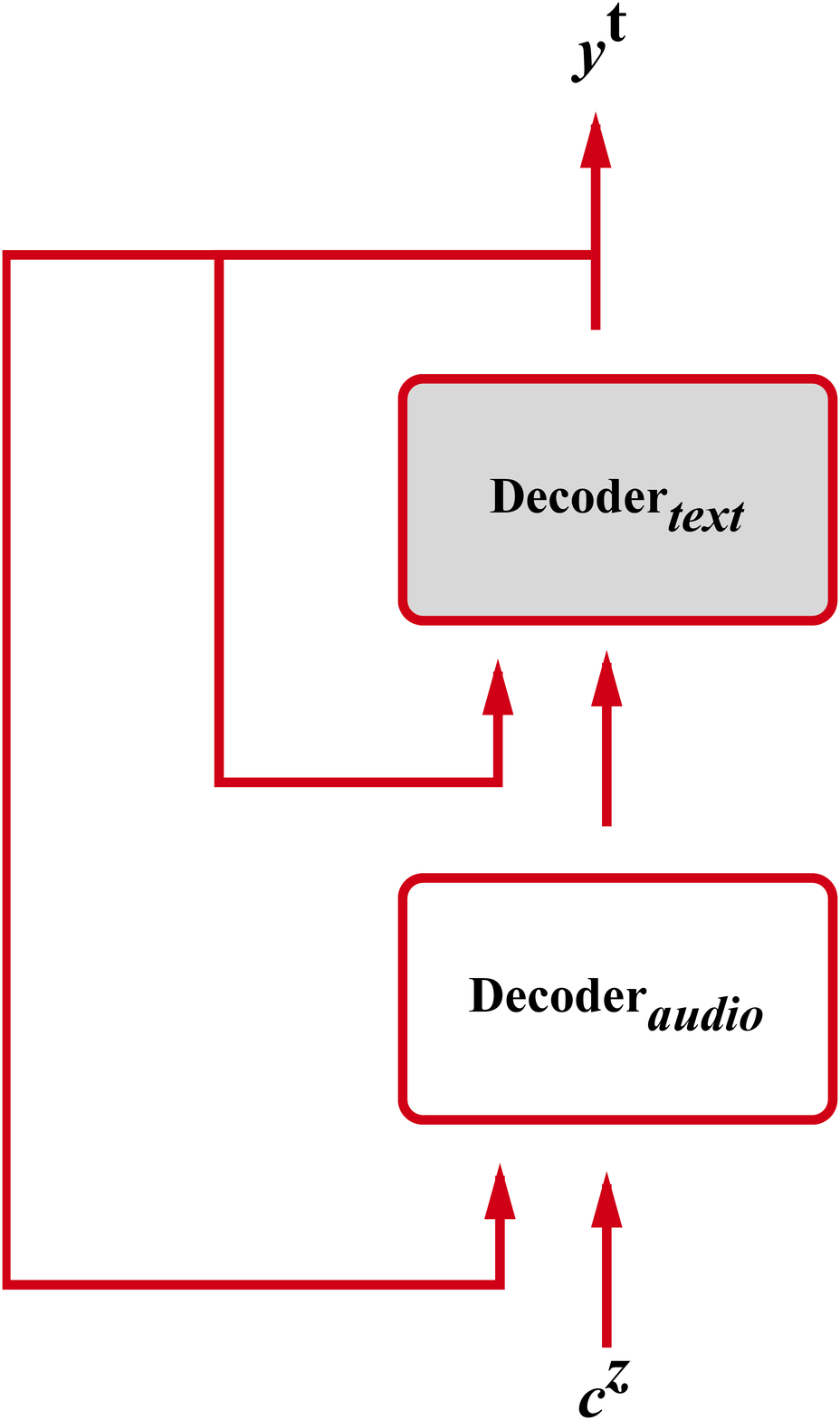}
        \caption{MUTE-ZT}
        \label{fig:mutes-c}
    \end{subfigure}%
    \begin{subfigure}[b]{.22\textwidth}
        \includegraphics[width=.6\linewidth]{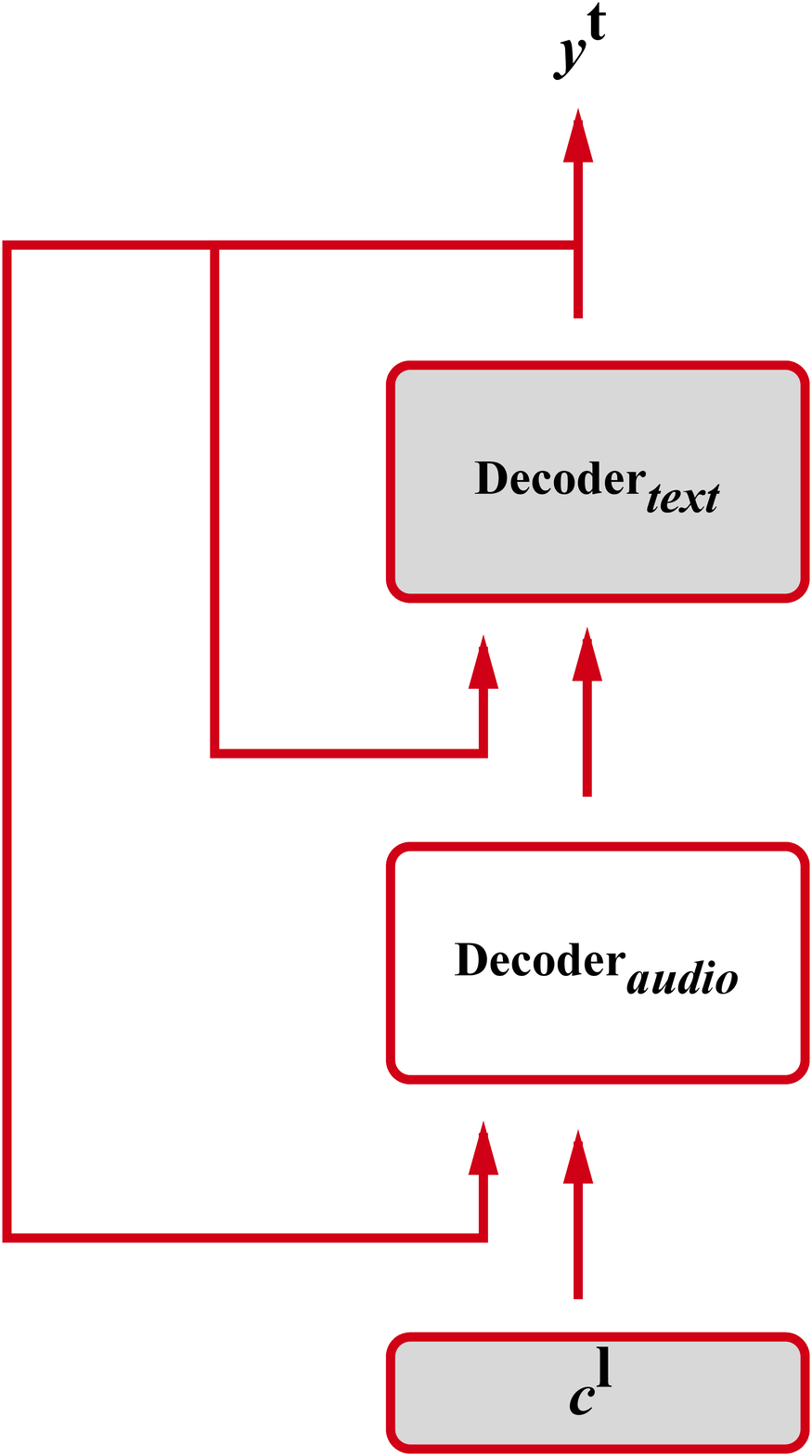}
        \caption{MUTE-LT}
        \label{fig:mutes-d}
    \end{subfigure}%
\centering
 \caption{Illustration for the four variations of MUTE decoders at training stage 2 using text-only data. See Fig. \ref{fig:mute_overview} caption for the meaning of different colors.}
 \label{fig:mutes}
\end{figure*}

\begin{figure*}[ht]
  \centering
  \includegraphics[width=0.5\linewidth] 
  {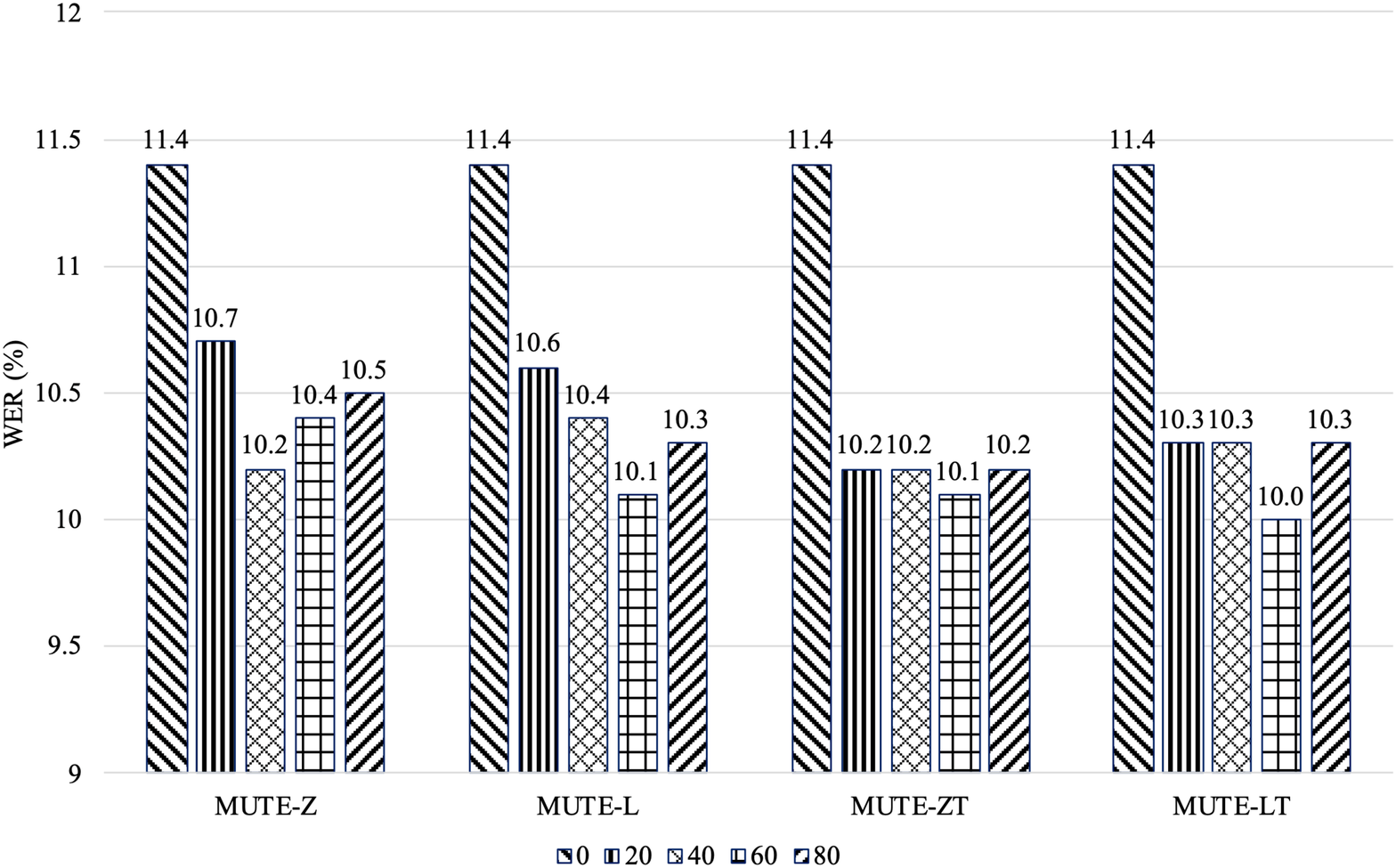}
\centering
 \caption{WERs of MUTE variations on LibriSpeech test-clean. Each bar corresponds to a different text-only data mixing ratio.} 
 \label{fig:txsr_comp}
\end{figure*}

\subsection{Multitask Training with Text Data}
\label{ssec:txsr}
Fig. \ref{fig:mute_overview} shows the two stages of MUTE training. In the first stage, the whole model is trained using the available audio-text pairs. In the second stage, we retrain the decoder by alternating training steps on audio-text pairs and text-only examples.
%
With audio-text pairs as training data, expression (\ref{eq:conventional}) can be simplified to:
\begin{equation}
    \label{eq:conv_dec}
    p({y}^{a}_u | {c}_u^{a}, \boldsymbol{y}^{a}_{1:u-1}; \theta_{d})
\end{equation}
where ${c}_u^{a} \in \mathbb{R}^{U^{a} \times E}$ refers to the context vector extracted from the encoder embedding for step $u$. $E$ denotes the dimension of the context vector and $\theta_{d}$ denotes the parameters in the decoder.
For text-only examples, we denote the output token sequence as $y^{t} \in \mathbb{R}^{U^{t}}$: 
\begin{equation}
    \label{eq:txsr_dec}
    p({y}^{t}_u | {c}^{t}, \boldsymbol{y}^{t}_{1:u-1}; \theta_{d})
\end{equation}
where $c^t$ corresponds to a specialized context vector, used for all output steps, indicating to the decoder layers that there is no audio context for this example and it should instead rely only on its internal state and the auto-regressive input to predict the next token.

Steps taken on text-only data help expose the model to a larger corpus of text data during training. Since the decoder parameters are mostly shared with the ASR task, this next-token prediction LM task prevents the decoder from over-fitting the small amount of audio-text pairs.
Meanwhile, steps using audio-text data ensures that the decoder can perform ASR tasks when conditioned on audio features.
By sweeping over the mixing ratios of audio-text and text-only data, MUTE takes the best from both ASR and LM tasks.

\subsection{Model Variants}
\label{ssec:txsr_variants}
We design the model architecture of MUTE from two perspectives, the choice of context vector $c^{t}$ in equation (\ref{eq:txsr_dec}) and the feedback loop for text-only data. 


The context vector ${c}^{t}$ in equation (\ref{eq:txsr_dec}) can be of two types, constant or learnable. We set the values of the constant context vector to zeros. The learnable context vector is shared for different frames and its weights are learned using the ASR objective function. The zero and learnable context vectors are denoted as $c^{z}$ and $c^{l}$, respectively. The corresponding models are referred to as MUTE-Z and MUTE-L, and are shown as Fig.~\ref{fig:mutes}-(a) and Fig.~\ref{fig:mutes}-(b). In MUTE-Z and MUTE-L, the feedback loop is shared for audio-text pairs and text-only data. For MUTE-ZT and MUTE-LT, we use an additional feedback loop specifically for text-only data. For audio-text training samples, MUTE-ZT and MUTE-LT update the entire decoder, whereas for text-only data, the two models only update the text-only loop. The two architectures are depicted in Fig. \ref{fig:mutes}-(c) and Fig. \ref{fig:mutes}-(d), respectively. Note that the text-only loop is inside the audio-text loop so that the whole decoder can take audio-text pairs during inference.









\section{Experimental Setup}
\label{sec:exp}

\subsection{Data and Model}
\label{ssec:data_model}
We conduct experiments on the LibriSpeech corpus. The training data for MUTE is a mixture of audio-text pairs and text-only data. For most experiments, the audio-text pairs are from the train-clean-100 subset and the text-only data is from the LibriSpeech-LM corpus, which contains about 40 million sentences. We also conduct experiment on the whole 960-hour LibriSpeech training set. The experiments on LibriSpeech 100h may better simulate a realistic low resource condition, where the amount of text-only data is much larger than that of transcribed audio-text pairs.
We vary the mixing ratio of text-only data in the whole training set from 0\% (i.e. the baseline) to 80\%, with a stride of 20\%.  We use the standard evaluation sets in the LibriSpeech corpus.

The E2E models in our experiments are 8-layer listen, attend, and spell (LAS) models based on the large model from \cite{irie2019choice}. The encoder consists of 2 batch-normalized convolutional layers, followed by 4 bidirectional long short-term memory (LSTM) layers. The decoder has 4 unidirectional LSTM layers. All LSTM layers contain 1024 nodes.
For MUTE-ZT and MUTE-LT, we use the top two layers in the decoder as $\textrm{Decoder}_{text}$ and the remaining two layers as $\textrm{Decoder}_{audio}$. We also use a language model trained on the LibriSpeech-LM corpus for the comparison between MUTE and shallow fusion. The LM is comprised of 2 LSTM layers, each consisting of 2048 nodes. Note that we use LAS models for their simplicity in incorporating text-only data. We do not perform data augmentation to exclude the influence of distorted audio features to our analysis. Note that JATD cannot work in our experimental setup since it requires unlabelled audio features and an external ASR model. 
We thus do not report the results of JATD.


\subsection{Implementation Details}
\label{ssec:details}
In the first training stage, we train the entire LAS model using audio-text pairs. In the second stage, we fix MUTE encoder parameters to those found during the first stage and randomly re-initialize the decoder. In initial experiments we found similar performance when initializing the decoder parameters using the first stage parameters. It is important to note that the encoder batch normalization layer statistics need to be fixed to ensure convergence.
The decoders are trained using a mixture of two types of data, audio-text and text-only. We randomly pick one type of sample at each training step with the mixing ratio described in Section~\ref{ssec:data_model}. Within each batch all samples are of the same type. To minimize the influence of hyperparameter selection, we use exponential moving average with a constant learning rate of $10^{-3}$ to train all the baseline and MUTE models. All models are trained for extensive number of epochs. The best models on the validation set are used for evaluation.

\section{Evaluation Results}
\label{sec:eval}

\subsection{Model Variant Selection}
\label{ssec:eval_arc}
We first compare and select one variant of MUTE for further experiments. For each of the model variants, we need to compare the results on different mixing ratios of the text-only and audio-text data. Fig. \ref{fig:txsr_comp} shows the test-clean WERs of the four types of MUTEs with various mixing ratios. MUTE-Z performs the best with 40\% text-only data, and all other MUTEs with 60\%.

Table \ref{tab:txsrs_comp} shows the WER comparisons among the MUTEs with their optimal training data mixing ratios. Although MUTE-LT performs the best on test-clean, MUTE-L is comparable to MUTE-LT on test-clean, performs better on test-other, and requires less modification to the decoder architecture. We thus experiment with MUTE-L for the remainder of this paper. Note that the models with the text-only feedback loop perform worse on test-other than those without it. The reason may be that $\textrm{Decoder}_{text}$ is not directly exposed to the audio features during training.

\begin{table}[t]
    \setlength{\tabcolsep}{3pt}
    \centering
    \caption{WER comparisons among MUTEs.}
    \label{tab:txsrs_comp}
    \begin{tabular}{l c c c c}
        \toprule
        Model & dev-clean & dev-other & test-clean & test-other \\
        \midrule
        MUTE-Z & 9.8 & 29.6 & 10.2 & 31.1 \\
        MUTE-L & 9.7 & 29.4 & 10.1 & 30.4 \\
        MUTE-ZT & 9.6 & 29.4 & 10.1 & 31.4 \\
        MUTE-LT & 9.6 & 29.5 & 10.0 & 30.9 \\
        \bottomrule
    \end{tabular}
\end{table}

\begin{table*}[htbp]
    \setlength{\tabcolsep}{1ex}
    \centering
    \caption{WER and oracle WER comparisons between the baseline and MUTE trained using LibriSpeech 100h and 960h. The models decoded using LM shallow fusion are denoted as those with \emph{+ LM}.}
    \label{tab:baseline_txsr}
    \begin{tabular}{l c c c c c c c c c}
        \toprule
        \multirow{2.5}{*}{Model} & \multirow{2.5}{*}{Train Set} & \multicolumn{4}{c}{WER} & \multicolumn{4}{c}{Oracle WER} \\
        \cmidrule(lr){3-6}  \cmidrule(lr){7-10}
        & & dev-clean & dev-other & test-clean & test-other & dev-clean & dev-other & test-clean & test-other \\
        \midrule
        Baseline & 100h & 10.8 & 30.8 & 11.4 & 32.2 & 7.7 & 26.6 & 8.1 & 28.0 \\
        MUTE & 100h & 9.7 & 29.4 & 10.1 & 30.4 & 6.8 & 25.0 & 7.3 & 26.2 \\
        \addlinespace
        Baseline + LM & 100h & 8.8 & 27.6 & 9.7 & 28.8 & 7.0 & 24.7 & 7.7 & 25.8 \\
        MUTE + LM & 100h & 8.6 &  27.4 & 9.3 & 28.4 & 6.5 & 23.7 & 7.1 & 25.1 \\
        \midrule
        Baseline & 960h & 4.5 & 13.6 & 4.7 & 13.7 & 2.4 & 9.7 & 2.6 & 9.5 \\
        MUTE & 960h & 4.1 & 11.9 & 4.2 & 12.1 & 2.1 & 8.2 & 2.2 & 8.0 \\
        \addlinespace
        Baseline + LM & 960h & 3.3 & 10.3 & 3.6 & 10.3 & 2.0 & 8.1 & 2.4 & 7.8 \\
        MUTE + LM & 960h & 3.4 & 10.3 & 3.6 & 10.3 & 1.9 & 7.7 & 1.9 & 7.1 \\
        \bottomrule
    \end{tabular}
\end{table*}

\subsection{WER and Oracle WER Comparisons}
\label{ssec:eval_txsr}
The top of Table \ref{tab:baseline_txsr} compares baseline and MUTE models trained on LibriSpeech 100h in terms of WER and oracle WER (the minimum WER in the n-best decoding list). For WER, MUTE outperforms the baseline by 11\% relatively on test-clean. When both models are shallow fused with the external LM, the relative improvement is 4\%. Note that our comparison is mainly on test-clean since we use the clean data (i.e. LibriSpeech 100h) during training. For oracle WER, MUTE without shallow fusion performs better than the baseline with shallow fusion, indicating that the best hypothesis in the beam is improved with the text-only data.


The bottom of Table \ref{tab:baseline_txsr} compares models trained on the whole LibriSpeech 960h training set. For WER, MUTE achieves the same relative improvement 11\% over the baseline on test-clean and a relative improvement of 12\% on test-other. The similar improvement to LibriSpeech 100h dataset demonstrates that MUTE is still helpful in a large training corpus setting. 
For oracle WER, MUTE alone outperforms the baseline using LM shallow fusion on test-clean, which is also consistent with the results on LibriSpeech 100h.


\subsection{Error Analysis} 
\label{ssec:eval_analysis}
We analyze MUTE by comparing different types of errors and sample output sentences using the models trained on LibriSpeech 100h.

\begin{table}[thbp]
    \setlength{\tabcolsep}{1ex}
    \centering
    \caption{Comparisons of different types of errors for models trained on LibriSpeech 100h. The results are shown in the order of deletion/insertion/substitution. See Table \ref{tab:baseline_txsr} caption for acronyms.}
    \label{tab:wer_types}
    \resizebox{\columnwidth}{!}{
    \begin{tabular}{l c c c c}
        \toprule
        Model & dev-clean & dev-other & test-clean & test-other \\
        \midrule
        Baseline        & 1.0/1.4/8.4 & 3.2/3.5/24.1 & 1.2/1.7/8.6 & 3.3/3.7/25.2 \\
        MUTE            & 1.3/1.1/7.2 & 3.9/2.9/22.6 & 1.4/1.3/7.4 & 3.9/3.0/23.5 \\
        Baseline + LM   & 1.6/1.0/6.2 & 4.9/3.1/19.6 & 2.1/1.1/6.5 & 5.3/3.3/20.2 \\
        MUTE + LM       & 1.5/1.0/6.2 & 4.5/2.7/20.2 & 1.9/1.0/6.4 & 4.5/2.8/21.2 \\
        \bottomrule
    \end{tabular}
    }
\end{table}

\begin{table}[!t]
    \setlength{\tabcolsep}{2pt}
    \centering
    \caption{Sample output sentences on test-clean. In the first row of each pairs of samples, wrong words are highlighted in red, whereas in the second row, the corresponding correct words are highlighted in green.}
    \label{tab:output_sentences}
    \begin{tabular}{@{}l p{0.74\linewidth}@{}}
        \toprule
        Model & Sentence \\
        \midrule
        {Baseline} & i haven't had a chance \colorbox{red!50}{get} to tell you what a jolly little place i think this is \\
        {MUTE} & i haven't had a chance \colorbox{green!50}{yet} to tell you what a jolly little place i think this is \\
        \midrule
        Baseline + LM & each of us is \colorbox{red!50}{lash} to some part of the raft \\
        MUTE + LM & each of us is \colorbox{green!50}{lashed} to some part of the raft \\
        \midrule
        {Baseline} & milligram roughly one twenty eight thousand of an \colorbox{red!50}{house} \\
        {MUTE} & milligram roughly one twenty eight thousand of an \colorbox{green!50}{ounce} \\
        \midrule
        {Baseline + LM} & \colorbox{red!50}{madame} roughly one twenty eight thousand of an \colorbox{red!50}{house} \\
        {MUTE + LM} & \colorbox{green!50}{milligram} roughly one twenty eight thousand of an \colorbox{green!50}{ounce} \\
        \bottomrule
    \end{tabular}
\end{table}

Table \ref{tab:wer_types} shows the deletion, insertion, and substitution errors of different methods. The results of MUTE and LM shallow fusion have the same trend: deletion errors increase, insertion errors decrease, and substitution errors decrease. This indicates that MUTE has a similar impact to LM shallow fusion on the error distribution. In Table \ref{tab:output_sentences}, we analyze the ability of MUTE to incorporate language level information by comparing four pairs of sample output sentences on test-clean. Since the goal of this analysis is to understand the sorts of errors that are not made when using MUTE training, the samples are selected such that MUTE contains less errors than the baseline. In the first pair of sentences, the baseline generates ``get'' and MUTE outputs ``yet''. These two words are similar in pronunciation but ``a chance yet to'' is better grammatically. In the second pair, ``is lashed to'' generated by MUTE uses the correct tense. In the following two output sentence pairs, we show the results of a single utterance using different methods. The ``ounce'' generated by MUTE fits better into the context ``milligram'' and ``an''. In addition to the comparison between MUTE and the baseline, we can observe from the table that baseline + LM may contain more errors than the baseline alone. This indicates that with the language information limited to audio-text pairs, the baseline ASR model could generate incorrect outputs that mislead LM shallow fusion.

\section{Concluding Remarks}
\label{sec:conc}
We have proposed MUTE, a two-stage multitask training approach for attention-based end-to-end speech recognition models to incorporate language level information. Text-only data is used to regularize the training of the decoder in a multitask manner. Trained using LibriSpeech 100h as audio-text data, MUTE outperforms the baseline by 11\% relatively on the test-clean evaluation set. It approaches the performance of shallow fusion and does not need the additional LM. We observe a similar trend on the LibriSpeech 960h training set. Analyses of different types of errors and sample output sentences show that MUTE can incorporate language level information effectively. Future work includes designing test-time training/adaptation methods for MUTE, combining MUTE with audio-only techniques, expanding MUTE to transducer based models, and applying MUTE to deliberation tasks.


\bibliographystyle{IEEEtran}
\bibliography{refs}

\end{document}